\documentclass[11pt,a4paper]{article}
\usepackage[dvipsnames]{xcolor}
\usepackage[hyperref]{emnlp2020}
\usepackage{graphicx}
\usepackage{times}
\usepackage{pifont}
\usepackage{latexsym}

\usepackage{booktabs}
\usepackage{microtype}
\usepackage{xspace}
\usepackage{amsmath}
\usepackage{amssymb}
\usepackage{soul}
\usepackage{multicol,multirow}

\newcommand\hmm[1]{\ifnum\ifhmode\spacefactor\else2000\fi>1000 \uppercase{#1}\else#1\fi}
\newcommand{\taskname}{\hmm{C}ontrollable Debiasing\xspace}
\newcommand{\modelname}{\textsc{PowerTransformer}\xspace}
\newcommand{\modelshort}{\textsc{PowerT}}

\def\aclpaperid{2072} 

\aclfinalcopy 

\usepackage[T1]{fontenc}
\usepackage{inconsolata}

\title{
    \modelname: \\Unsupervised Controllable Revision for Biased Language Correction
}
\newcommand\coauth{$^\star$~}

\newcommand{\aiTwo}{$^\dagger$}
\newcommand{\uw}{$^\diamond$}
\newcommand{\aspace}{\hspace{1em}}

\author{
Xinyao (Michelle) Ma\coauth\uw \aspace
    Maarten Sap\coauth\uw \aspace
    Hannah Rashkin\uw \aspace
    Yejin Choi\uw\aiTwo \\
\uw Paul G. Allen School of Computer Science \& Engineering, University of Washington\\
\aiTwo Allen Institute for Artificial Intelligence\\
Seattle, USA\\
\texttt{\{max36,msap,hrashkin,yejin\}@cs.washington.edu}
}

\begin{document}

\maketitle
\begin{abstract}
    Unconscious biases continue to be prevalent in modern text and media, calling for algorithms that can assist writers with bias correction. For example, a female character in a story is often portrayed as passive and powerless (``\textit{She daydreams about being a doctor}'') while a man is portrayed as more proactive and powerful (``\textit{He pursues his dream of being a doctor}''). 

We formulate \textbf{\taskname}, a new revision task that aims to rewrite a given text to correct the implicit and potentially undesirable bias in character portrayals. 
%
%
We then introduce \modelname as an approach that debiases text through the lens of \emph{connotation frames} \cite{sap2017connotation}, 
which encode pragmatic knowledge of implied power dynamics with respect to verb predicates.
One key challenge of our task is the lack of parallel corpora.
To address this challenge, we adopt an unsupervised approach using auxiliary supervision with related tasks such as paraphrasing and self-supervision based on a reconstruction loss, building on pretrained language models.
%
%

Through comprehensive experiments based on automatic and human evaluations, we demonstrate that our approach outperforms ablations and existing methods from related tasks.
Furthermore, we demonstrate the use of \modelname as a step toward mitigating the well-documented gender bias in character portrayal in movie scripts. 
    \let\thefootnote\relax\footnotetext{\coauth Both authors contributed equally.}
\end{abstract}

\section{Introduction}
\begin{figure}[t]
    \centering
    \includegraphics[width=\columnwidth]{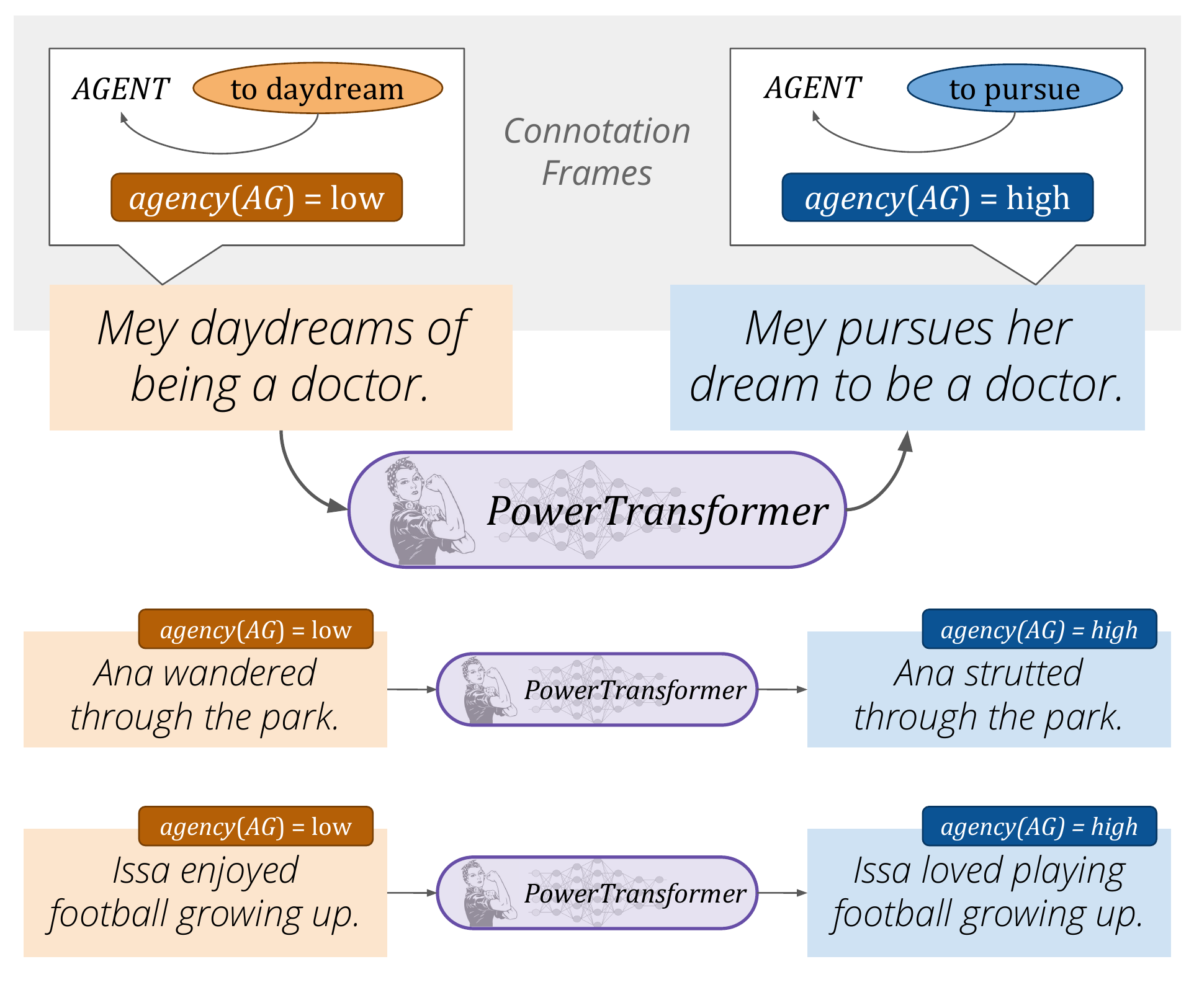}
    \caption{Examples of using connotation frames \cite{sap2017connotation} for controllable revisions to portray characters with more agency and power.
    In the second example, ``Ana strutted'' implies that she is more active and decisive, compared to ``Ana wandered'' which portrays her as aimless and passive.
    }
    \label{fig:introFig}
\end{figure}

Narratives and news texts often reflect societal biases and stereotypes, such as the traditional gender role that women are passive and submissive \cite{lakoff1973language,Fiske1993controlling,fast2016shirtless}.
The task of \textit{controllable text revision}, i.e., rephrasing text to a targeted style or framing, can help correct for these biases by altering and equalizing the way people are described.
For example, automatically rewriting \emph{``Mey daydreamed about being a doctor''} as \emph{``Mey pursued her dream to be a doctor''} portrays Mey with more authority and decisiveness (Figure \ref{fig:introFig}). 
Such controllable revision methods could be used to help reshape how gender roles are portrayed in media \cite[e.g., through machine-in-the-loop writing systems;][]{clark2018creative}.

To edit such biases out of text, a controllable rewriting model faces three key challenges.
First, a model should be able to make edits beyond surface-level paraphrasing, as simple paraphrasing will often not adequately debias the underlying events described.
For example, Mey's portrayal in Figure \ref{fig:introFig} carries both overt bias (the choice of action) and subtle bias (the framing of the action), both of which require rewriting to be adequately debiased.
Second, a model's debiasing revisions should be purposeful and precise and should not make unnecessary changes to the underlying meaning of the original text.
Lastly, since parallel data does not exist, models must learn to revise and debias text without supervised data, thereby preventing straightforward machine translation-style modelling.

We formulate \textit{\taskname} as a new controllable text revision task that 
aims to correct the implicit and possibly unwanted bias against or towards a specific character portrayed in text (\S\ref{sec:task-desc}). 
As shown in Figure \ref{fig:introFig} (top), we 
study the portrayal biases
through the lens of connotation frames of \textit{power and agency} \cite{sap2017connotation}, which provide pragmatic knowledge about implied power and agency levels projected onto characters by a predicate.

We create \modelname, an encoder-decoder model that rewrites sentences with a desired portrayal using 
agency connotation frames (\S\ref{sec:model-desc}).
We combine a reconstruction and paraphrase objective into our model to overcome the lack of parallel supervised data, building off of the denoising autoencoder setup from \citet{Li2018-op}.
To steer the revisions, we endow the model with connotation frame knowledge both at training time using control tokens, and at generation time using agency-based vocab boosting.

Our findings show that \modelname is effective at rewriting sentences with desired agency connotations while only making minimal changes to their meaning, as measured through both human and automatic evaluations (\S\ref{sec:experiments}).
We also show that \modelname significantly outperforms existing stylistic rewriting methods \cite{Prabhumoye2018-kj,Dathathri2020-ua} on those metrics. 
Additionally, through ablations studies, we establish the usefulness of each component of the model, finding benefits from both the joint objective (47\% gain in accuracy) and the agency scaling (12\% gain in accuracy).


Finally, in \S\ref{sec:movies}, we apply \taskname to a corpus of modern English movies \cite{gorinski2015movie} as a step towards removing gender bias in character portrayal established by prior work \cite{sap2017connotation}.
Using \modelname, we revise the movie scripts and significantly increase the agency levels of female characters, thereby reducing the 
gender bias.
Our findings show promise for using modern NLP tools to help mitigate societal biases in text.
We release our preprocessed data and code at \url{http://maartensap.com/controllable-debiasing}.







\section{\taskname}
\label{sec:task-desc}


\taskname is a novel formalization of stylistic rewriting that aims to debias the portrayal of characters through controllable revision. 
To achieve the desired character portrayal, a system must be able to change the underlying meaning of events, unlike certain formalizations \cite[e.g., politeness transfer;][]{Rao2018-fd} where full meaning preservation is required.
Without this, systems run the risk of merely paraphrasing the biases in text.
However, revisions must be precise and avoid unnecessary meaning changes, which can often occur in stylistic rewriting (e.g., reversing the sentiment of a review drastically changes its underlying meaning).

For our new rewriting task of changing portrayal bias,
we focus on connotation frames that measure the \textit{power} and \textit{agency} ascribed to characters through the actions they take.
Connotation frames \cite{rashkin2016connotationframes,sap2017connotation} distill implicit relations between a verb, its agent, and its theme.
In this work, we use the positive, neutral, and negative agency dimensions, where agency is defined as the capacity to intentionally make changes or act upon one's environment \cite{dennett1989intentional}.
For example, illustrated in Figure \ref{fig:introFig}, ``X pursued Y'' implies that X has positive agency.\footnote{Future work could explore using the power dimension instead of agency, or alternative operationalizations of biases, e.g., Social Bias Frames \citep[][]{sap2020socialbiasframes} or \textit{regard} towards minorities as introduced by \citet{sheng-etal-2019-woman}.}
Using machine-in-the-loop writing systems \cite[e.g., ][Textio\footnote{\url{https://textio.com/}}]{ghazvininejad2016generating,ghazvininejad2017hafez,clark2018creative},
models trained on this task could help authors write news, stories, or movies that portray characters in less biased ways, and thereby help mitigate the negative effects of stereotypical portrayals in media \cite{behm2008mean,field2019contextual}.


\section{\modelname}

\label{sec:model-desc}

\newcommand{\joint}{{\it Joint}}
\newcommand{\paraOnly}{{\it ParaOnly}}
\newcommand{\boost}{{\it Boost}}
\newcommand{\noBoost}{{\it noBoost}}
\newcommand{\supplyVerb}{{\it SupplyVerb}}
\newcommand{\seqToSeq}{seq2seq w/ transformer}
\newcommand{\bst}{BST\xspace}
\newcommand{\bstCite}{\bst \cite{Prabhumoye2018-kj}}
\newcommand{\pplm}{PPLM\xspace}
\newcommand{\pplmCite}{\pplm \cite{Dathathri2020-ua}}


\begin{figure*}[t]
    \centering
    \includegraphics[width=.98\textwidth,page=1,clip,trim=0 11pt 0 11pt]{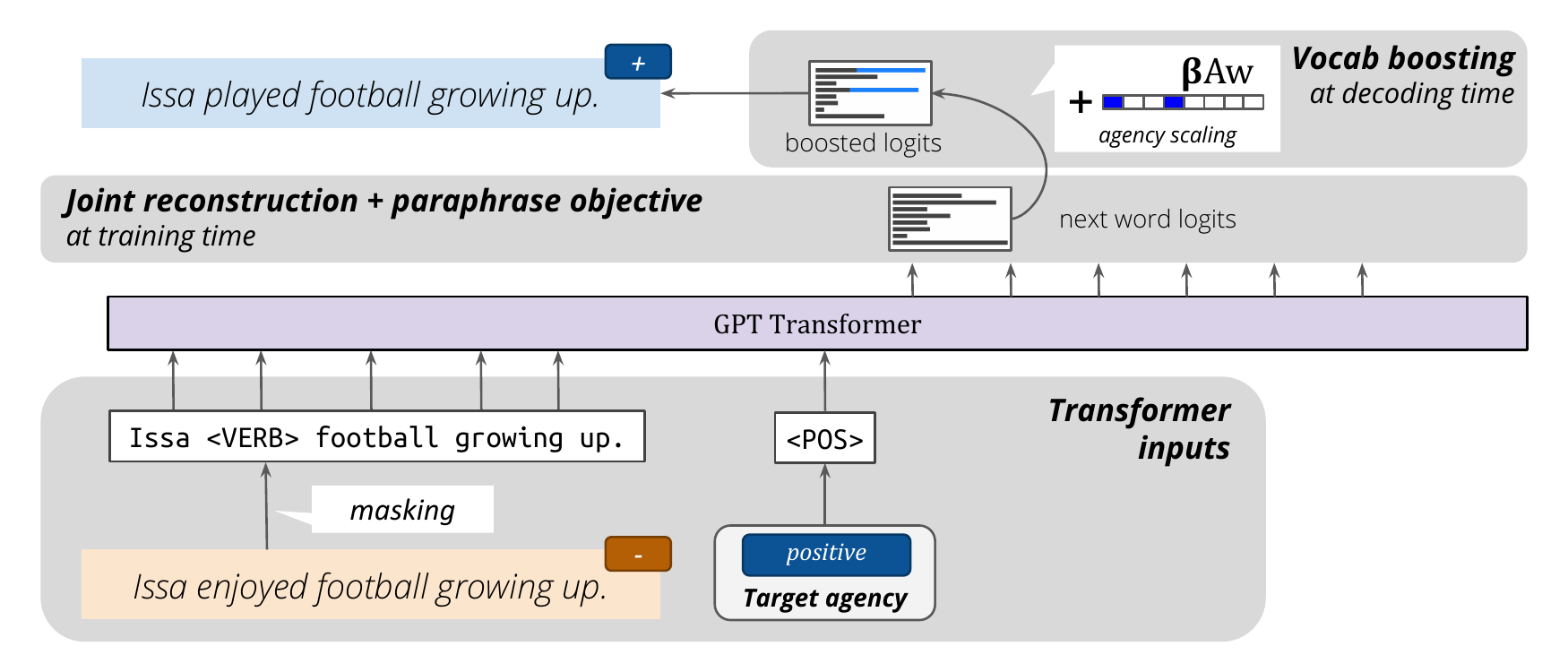}
    \caption{Overview of the full \modelname{} model.  An input sentence is masked for verb tokens indicative of agency.  Masked inputs and target agency are used as GPT inputs.  We use a joint objective using both paraphrase data and masked input sentences for training.  At decoding time, we employ a vocab boosting technique to steer generations towards the target agency.}
    \label{fig:modelfig}
\end{figure*}

We present a new approach for \taskname called \modelname, which addresses two key challenges: the paucity of parallel supervised data for training and the difficulty of incorporating fine-grained control for steering the agency of the output.
Our approach (Figure \ref{fig:modelfig}) jointly learns to reconstruct partially masked story sentences while also learning to paraphrase from an external corpus of paraphrases (\S\ref{ssec:joint-objective}).
At generation time, we also include a boosting method for fine-grained steering towards the desired agency level as described in \S\ref{ssec:logits}.

\subsection{Model Overview}
\modelname{} is an encoder-decoder style model with an OpenAI-GPT transformer model \cite{radford2018improving} as the base.
The input sentence $\mathbf{x}$ is converted to a sequence of byte pair encodings (BPE) $\{x_1,...,x_n\}$, and given to the encoder after being scrubbed of its agency markers
as described below. 
To steer the model, we also give the encoder the target agency $t$, which we represent as one of three special tokens \{\texttt{$<$Pos$>$},\texttt{$<$Equal$>$},\texttt{$<$Neg$>$}\}.%
\footnote{In earlier experiments, we also provided the original agency as an input to the model during training and decoding, but found that it made little difference in performance.}

\subsection{Joint Objective}\label{ssec:joint-objective}
We train our model 
on both a reconstruction and a paraphrasing task, for which inputs are masked and paraphrased versions of the output, respectively.
\begin{equation}\label{eq:objective}
    \mathcal{L}_\text{joint} = \mathcal{L}_\text{recon} + \mathcal{L}_\text{para}
\end{equation}


\paragraph{Masking and Reconstructing}
\label{ssec:denoise}
Inspired by the delete-retrieve-generate model from \citet{Li2018-op},
this objective teaches the model to recover masked out agency-associated verbs in sentences.
We first assign an agency level to an input sentence by counting verbs in the agency lexicon from \citet{sap2017connotation}.%
\footnote{For sentences that have multiple verbs, we assign the agency level that the most verbs in the sentence have (e.g., a sentence with two positive agency verbs and one negative agency verb will be assigned positive agency).} 
Then, we mask out all verbs indicative of the agency level,  
replacing them with a special \texttt{$<$VERB$>$} token. 
In this setup, the target output is the original sentence $\mathbf{x}=\{x_1,...,x_n\}$, with the masked sentence $\mathbf{\hat{x}}$ and the target agency level $t$ as inputs.
During training, we minimize the cross entropy of the target output sentence given the inputs:
\begin{equation}\label{eq:obj_recon}
     \mathcal{L}_\text{recon} = - \frac{1}{n}\sum_{i=1}^n \log p (x_i | x_{<i}, \mathbf{\hat{x}}, t)
\end{equation}

\paragraph{Paraphrasing}
To go beyond reconstructing sentences, we add a paraphrasing objective
using an out-of-domain paraphrase corpus (\S\ref{ssec:datasets}).
We extract agency levels for each sentence and its paraphrase and mask out the agency verbs in the input, using the same methods as described above.
Here, the inputs are the masked sentence $\mathbf{\hat{x}}$ and the target agency $t$, while the target output $\mathbf{y}=\{y_1,...,y_m\}$ is the paraphrase.
As with reconstruction, we minimize the cross entropy of the target output given the inputs:
\begin{equation}\label{eq:obj_para}
     \mathcal{L}_\text{para} = - \frac{1}{m}\sum_{i=1}^m \log p (y_i | y_{<i}, \mathbf{\hat{x}}, t)
\end{equation}

\subsection{Controlled Decoding with Vocab Boosting}
\label{ssec:logits}
\begingroup
We employ a vocab-boosting technique during generation to encourage models towards generating with the desired agency, inspired by   \citet{ghosh2017affectlm}. 
At each decoding timestep $i$, we re-scale the unnormalized token probabilities (logits $l_i \in \mathbb{R}^{V}$, where V is the vocabulary size) to boost the likelihood of predicting words with the target agency.
The next token probabilities are then computed using the ``boosted'' logits:
\begin{equation}
    P(y_i | y_{<i}, x, t) \propto \text{softmax}(l_i + \beta \cdot A  w)
\end{equation}
where $A$ is a $\mathbb{R}^{V\times3}$ matrix that represents a 3-dimensional \{positive, equal, and negative\} agency embedding for each token in the vocabulary, $w$ is a $\mathbb{R}^{3}$ one-hot vector denoting the target agency for the output, and $\beta$ is a scalar hyperparameter representing the boosting strength.
We create $A$ manually using the verbs in the agency lexicon \cite{sap2017connotation}.\footnote{Since our model operates on BPE tokens, we manually set the first BPE token of every tense of every verb to the desired agency.
We also experimented with learning $A$ from data, but found no improvement over manually setting it.
} 
Used only at decoding time, this method effectively increases the likelihood of using a word with the target agency level.

\endgroup



\section{\taskname Experiments}

\label{sec:experiments}

In this section, we describe three experiments for investigating \modelname performance.
First, we evaluate performance of our full model and ablated baselines, using automatic metrics to quantify the effectiveness of each modelling component (\S\ref{sec:ablation}).
Next, we compare our full model to baselines from related work (\S\ref{sec:externalbaselines}).
Lastly, given the limitations of automated metrics for evaluating generations \citep{HowNotToEval,Mir2019-bv}, we obtain
human judgments of model performance through crowdsourcing 
(\S\ref{sec:humaneval}).
We additionally include examples of generations in Table \ref{tab:examples}. 

\subsection{Datasets}
\label{ssec:datasets}

\begin{table}[t]
    \centering
    \begin{tabular}{@{}p{.2em}ccccc@{}}
    \toprule
         & Type & \# Instances & Pos & Neutral & Neg  \\ \midrule
         \multirow{3}{*}{\rotatebox{90}{\textit{ROC}}}& train & 10721 & 3834 & 4151 & 2736 \\
         & dev & 1803 & 633 & 710 & 460\\
         & test & 899 & 325 & 350 & 224\\ \midrule
         \multirow{2}{*}{\rotatebox{90}{\textit{Para.}}} & train & 45000  & 16410 & 14153 & 14437\\
         & dev & 10000 & 3645 & 3328 & 3127  \\
         \bottomrule
    \end{tabular}
    \caption{Statistics for our main story sentences dataset (ROC) and for the external paraphrase corpus (Para.).}
    \label{tab:data:stats}
\end{table}
In our experiments, we use a dataset of short stories for the reconstruction task and a parallel corpus of paraphrases for both paraphrase and reconstruction tasks.
We show data statistics in Table \ref{tab:data:stats}, with additional preprocessing details in Appendix \ref{sup:data-deets}.


\paragraph{ROC story corpus}
The main focus of our study is controllable revision of story sentences; therefore, we select sentences from the ROC story corpus \cite[ROC][]{Mostafazadeh2016ROCStory}.
After extracting agency levels for all sentences from the training stories, we sample roughly equal amounts of all three agency levels, and randomly split sentences into training, development, and test sets.\footnote{We use a 80:13:7 train, development, test ratio.}


\paragraph{Paraphrase corpus}
As additional training data, we use the corpus of automatically aligned paraphrases of TV subtitles \cite[][Para.]{Creutz2018opuparcus}.
As with the ROC story corpus, we extract agency levels for each sentence and its paraphrase, then sample roughly equal amounts of pairs with all different sentence-paraphrase agency combinations (further details in \S\ref{sup:para-deets}). 
We randomly split the data into 45k train and 10k dev. instances (Table \ref{tab:data:stats}).\footnote{Since this is just additional training data, we do not test our models on this corpus, but do use the dev. set for selecting some hyperparameters.}


\begin{table*}[t]
    \centering
    \begin{tabular}{lccccc}
    \multicolumn{6}{c}{\textbf{Ablations using the Development Set}}\\
    \toprule
        & \multicolumn{2}{c}{Main Metrics} & \multicolumn{3}{c}{Additional Metrics}\\
        \cmidrule(l{5pt}r{5pt}){2-3} \cmidrule(l{5pt}r{5pt}){4-6}
        & {\bf Agency} & {\bf Meaning} & {\bf Fluency} & {\bf Repetition} & {\bf Diversity}\\
         \modelname variants & Acc ($\uparrow$) & BertScore ($\uparrow$) & PPL ($\downarrow$) &  w/ Rep ($\downarrow$)& Unique ($\uparrow$)\\
         \midrule
        {(\paraOnly+\noBoost)}& .30 & .95 & \textbf{58.76} & .002 & .54\\
        {(\paraOnly+\boost)}& .42 & .90 & 76.25 & \textbf{.001} &.59\\ 
        
        {(\joint+\noBoost)} & .77 & \textbf{.96} &70.61&.007 &.87\\
       {(\joint+\noBoost)+\supplyVerb} & .77 & \textbf{.96} & 94.54 &.004&.92\\
       \midrule
        \textsc{Full}~=~{(\joint+\boost)} & \textbf{.89} & \textbf{.96} &76.78&.015&\textbf{.99}\\ 
    \bottomrule
    \end{tabular}
    \caption{Ablation study results on the development set.
    We present separate metrics for evaluating the change in agency, the meaning preservation, fluency, repetitiveness and diversity of the output (bolding the best performance).
    ($\uparrow$) indicates that higher is better and ($\downarrow$) indicates that lower is better.
    }
    \label{tab:results}
\end{table*}




\subsection{Metrics}
In addition to human evaluations, we also use a variety of automated evaluation metrics to characterize different aspects of performance. 
We measure the accuracy of the change in agency by comparing the 
target agency level with that of the output (extracted using the connotation frames lexicon).
As a measure of meaning preservation, we use BERT-score F1 metrics \citep{bertscoreppr} to compare the semantic similarity of the input sentence with the machine output.

As additional metrics, we measure the fluency, the repetitiveness, and diversity of the output.
Following previous work \cite{Dai2019-uw}, we measure fluency as perplexity (\textit{PPL}) of the output sentence using a pre-trained GPT model that has not been fine-tuned for this task.
As an additional metric of potential text degeneration, we compute
the fraction of output sentences that have a bigram that is repeated two or more times (\textit{w/ rep}).
Finally, 
we compute the fraction of generations that are unique with respect to the rest of the output, to ensure diverse, input-specific generations (\textit{unique}).

\subsection{Experimental Setup}
We randomize ROC story and paraphrase data, and use OpenAI GPT LM as our pretrained model. 
For decoding, we use top-$p$=0.4 nucleus sampling \cite{nucleussampling}, and a boosting strength of $\beta$=5 (hyperparameters and details in \S\ref{sup:model-hyperparams}).

\subsection{Investigating Effectiveness of Approach}
\label{sec:ablation}
We first establish our model's effectiveness at \taskname on our dev. set, and investigate the importance of various components in our approach through ablation analyses.
For qualitative analyses, we also show example revisions in Table~\ref{tab:examples} (and Table \ref{tab:more-examples} in the appendix).

\subsubsection{Ablated Baselines}
We first investigate the importance of the reconstruction objective,
by comparing our joint objective model (\joint) with a model trained with just the paraphrasing objective (without masking, \paraOnly).
Then, to quantify the effect of boosting, 
we compare models with (\boost) and without (\noBoost) agency-specific vocab boosting. 
Note that \paraOnly+\noBoost{} is equivalent to a GPT-based encoder-decoder model, similar to seq2seq frameworks commonly used in paraphrasing tasks \citep{seq2seqpara,li2018paraphrase,prakash-paraphrase}.

As a final comparison, we implement a model variant that more closely mirrors the delete-retrieve-generate paradigm \citep{Li2018-op} by adding a ``retrieve'' step in which we concatenate transformer input with a verb retrieved from the verb agency lexicon that is most similar to the masked out verb (\textit{\supplyVerb}).%
\footnote{We retrieve a verb from the \citet{sap2017connotation} lexicon that has the target agency and is most similar to the masked out verb, where similarity is defined as cosine distance between word embeddings using GloVe 300-d embeddings \citep{glove}.} 

\begin{table*}[t]
    \centering
    \begin{tabular}{lccccc}
    \multicolumn{6}{c}{\textbf{Test Set Comparisons} (pos-to-neg and neg-to-pos set)}\\
        \toprule
        & \multicolumn{2}{c}{Main Metrics} & \multicolumn{3}{c}{Additional Metrics}\\
        \cmidrule(l{5pt}r{5pt}){2-3} \cmidrule(l{5pt}r{5pt}){4-6}
        & {\bf Agency} & {\bf Meaning} & {\bf Fluency} & {\bf Repetition} & {\bf Diversity}\\
          & Acc ($\uparrow$) & BertScore ($\uparrow$) & PPL ($\downarrow$) & w/ rep ($\downarrow$)& unique ($\uparrow$)\\
         \midrule
        \pplmCite &.13 &.95 &106.12&.053&\textbf{1.00}\\
        \bstCite & \textbf{.88} &.83 & \textbf{91.22} & .053 &0.79\\
        \modelname & .86 & \textbf{.96} & 95.19 & \textbf{.015} &\textbf{1.00} \\
    \bottomrule
    \end{tabular}
    \caption{Performance of different re-writing methods on the neg-to-pos and pos-to-neg subsets of the test set (bolding the best performance). 
    We evaluate the change in agency and the meaning preservation. As secondary metrics, we include fluency,  repetitiveness, and diversity of the output.}
    \label{tab:results:test}
\end{table*}

\subsubsection{Results}
In Table~\ref{tab:results}, our results show that the full model (\joint+\boost) yields text revisions with the most accurate target agency and the most meaning preservation. 
In general, we find that both the joint objective and vocab boosting (\boost) substantially increase the target agency accuracy, as also illustrated in examples (d) and (e) in Table \ref{tab:examples}.
However, unsurprisingly, vocab boosting also slightly lowers fluency, yielding higher perplexities than models' non-boosted counterparts. 
Our results also show that using the joint objective with boosting increases the diversity of output, but causes marginally more repetition of bigrams.

Counterintuitively, our ablations show that supplying a verb to the model as an explicit retrieval step (\supplyVerb) does not improve the agency or meaning metrics and actually hurts the fluency of the output (as measured by higher perplexities).
Upon qualitative investigation (Table \ref{tab:more-examples} in the appendix), the retrieved verb is often related to a different word sense of the masked verb, breaking the grammaticality of the sentence.

\subsection{Comparison with External Approaches}
\label{sec:externalbaselines}
To further validate our approach, we compare against two baselines from related style transfer and stylistic generation tasks. 
As these models were designed for binary style transfer, we only report our baseline and model results on the positive and negative agency portions of our data.  

\subsubsection{Baselines}

\paragraph{\bst} We compare to the backtranslation style transfer model from \citet{Prabhumoye2018-kj}.
This model first translates input sentences to a pivot language (preserving the meaning but losing language-specific style), then relies on style-specific decoder-translators for generating the output sentence. We include set-up details in \S\ref{sup:bst-details}.

\paragraph{\pplm} 
Recent work in controllable generation 
has introduced PPLM, a new plug-and-play technique with promising results for decoding stylistic text \citep{Dathathri2020-ua}.
This method operates on an underlying neural language model at decoding time. It uses backpropagation from a stylistic discriminator to update the past and present hidden representations to be more consistent with the targeted style or domain.
We adapt the approach to controllable revision by replacing the base language model with an autoencoder trained on a reconstruction objective, described in detail in \S\ref{sup:pplm-details}.


\subsubsection{Results}
We present results in Table~\ref{tab:results:test}.
Our experiments show that \modelname performs better than the baselines overall.
Specifically, while the \bst{}  revisions obtain slightly higher accuracy on the output agency levels, these revisions have the both the lowest diversity and meaning preservation, suggesting the model ignores the input (Table~\ref{tab:examples}). 
\pplm{} shows opposite trends, yielding the lowest accuracy with high meaning preservation and high diversity of generations.
Illustrated in Table~\ref{tab:examples}, this model often makes less purposeful and less concise alterations.


\subsection{Evaluating with Human Judgements}
\label{sec:humaneval}
To validate our automatic evaluations, we collect human judgments of the controllable revisions from several baselines and \modelname(\joint+\boost).


\newcommand{\low}[1]{{\color{BurntOrange}\textbf{#1}}}
\newcommand{\high}[1]{{\color{RoyalBlue}\textbf{#1}}}
\newcommand{\correct}{\small\color{gray}\checkmark}

\begin{table*}[t!]
    \centering
    \small
    \begin{tabular}{c@{\hspace{.5em}}c@{\hspace{.5em}}p{2.25cm}@{\hspace{.7em}}r@{\hspace{.7em}}p{7cm}@{\hspace{.1em}}c}
    \toprule
          \textbf{dir.} & & \textbf{Input} & \textbf{Model} &
          \textbf{Revised Sentence (out)} &
          \multicolumn{1}{@{}c@{\hspace{.4em}}}{\textbf{Agency(out)}}\\    
        \midrule
        \multirow{17}{*}{\rotatebox{90}{\large \textit{agency}~{\Large(\high{+}~$\rightarrow$~\low{--})}}} & \multirow{5}{*}{(a)} &  \multirow{5}{2.5cm}{After the party I headed home.}
        & PPLM & after the party my classmate, Kayla and the Tgirls of the Universe. & =  \\
        && & BST & please 's , i have a word of this . & \low{--}\\
        & & & \modelshort$_{Joint+NoBoost}$ & after the party i stayed home. & \low{--}\\
        & & & \modelshort$_{Joint+Boost}$ & after the party i stayed home. & \low{--}\\
         \cmidrule(l){2-6}
         & \multirow{7}{*}{(b)} & 
         \multirow{7}{2.25cm}{A friend asked me to watch her two year old child for a minute.} 
        & PPLM & a friend asked me to watch her two year old child for a minute. & \high{+} \\
        & & & BST & l didn 't have a word of this , you 're . & \low{--}\\ 
        & & & \modelshort$_{Joint+NoBoost}$ &a friend needed me to watch her two year old child for a minute. & \low{--}\\
        & & & \modelshort$_{Joint+Boost}$ & a friend {needed} me to watch her two year old child for a minute. &\low{--}\\
         \cmidrule(l){2-6}
         & \multirow{5}{*}{(c)} &
         \multirow{5}{2.25cm}{After filling in the data it looked quite sharp.}
         & PPLM & before filling the last question it it it it looked quite sharp. before filling the last question it it & \high{+}\\
         & & & BST & when the 't you want a word ? & \low{--}\\
         & & & \modelshort$_{Joint+NoBoost}$ & after analyzing in the data it looked quite sharp. & =\\
         & & & \modelshort$_{Joint+Boost}$ & after seeing in the data it seemed quite sharp. &\low{--}\\
         \midrule
         \multirow{12}{*}{\rotatebox{90}{\large \textit{agency}~{\Large(\low{--}~$\rightarrow$~\high{+})}}} & 
         \multirow{4}{*}{(d)} &
         \multirow{4}{2.25cm}{Allie was failing science class.}
         & PPLM & allie was failing science grade. & \low{--}\\
         & & & BST & do you want me ? & \high{+} \\
         & & & \modelshort$_{Joint+NoBoost}$ & allie was failing science class. & \low{--}\\
         & & & \modelshort$_{Joint+Boost}$ & allie was taking science class. & \high{+}\\
         \cmidrule(l){2-6}
         
        
        &\multirow{4}{*}{(e)} &  \multirow{4}{2.25cm}{Darla wanted a soft drink.} & PPLM & darla wants a hard hard drink. & \low{--}\\
        & & & BST & don 't take me a man . & \high{+}\\
        & & & \modelshort$_{Joint+NoBoost}$ & darla ordered a soft drink. & \high{+}\\
        & & & \modelshort$_{Joint+Boost}$ & darla ordered a soft drink. & \high{+}\\
        \cmidrule(l){2-6}

        &\multirow{4}{*}{(f)}& \multirow{4}{2.25cm}{Clint paused on the trail.}  & PPLM & clint was on the trail. & =\\
        & & & BST & don 't you want me , & \low{--}\\
        & & & \modelshort$_{Joint+NoBoost}$ & clint hiked on the trail. & =\\
        & & & \modelshort$_{Joint+Boost}$ & clint walked on the trail heading down. & \high{+}\\
       
    \bottomrule
    \end{tabular}
    \caption{Example sentences from our dev. set, along with their revisions from various models and the achieved agency levels (Agency(out)).
    Examples (a)-(c) should be rewritten from high to low agency, and (d)-(f) from low to high agency.
    Confirming our quantitative results in Tables \ref{tab:results} and \ref{tab:results:test}, \modelname (\joint+\boost) is the most effective at making purposeful and precise changes to the input sentences to alter their agency while minimally changing their meaning.
    Revisions from more models are listed in Table \ref{tab:more-examples} (in the appendix).
    }
    \label{tab:examples}
\end{table*}
\begin{figure}[t]
    \centering
    \includegraphics[trim={.2cm .5cm .1cm 0cm},clip,width=\columnwidth]{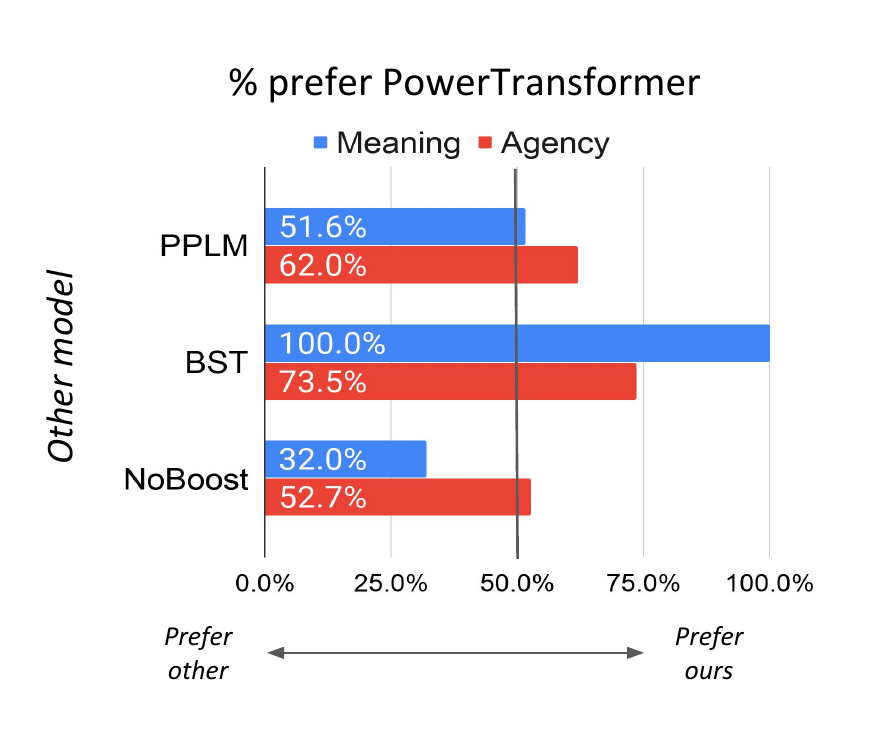}
    \caption{Human judgements of target agency and meaning preservation in \modelname{} vs. three other model variants.
    Selection rates $>$50\% indicate preference towards our model.
    }
    \label{fig:humaneval:H2H}
\end{figure}

\subsubsection{Human Evaluation Task}
We design a head-to-head%
    \footnote{We use head-to-head evaluations as those have been shown to be more reliable than scale-rating evaluations \cite{kiritchenko-mohammad-2017-best}.}
crowdsourcing task on Amazon Mechanical Turk where we ask raters to compare two outputs from different models given the same input sentence and target agency (see Figure \ref{fig:mturk-task} in the appendix).
We first ask them to judge whether either output is gibberish, then, in two questions, choose which revision has better targeted agency and which better preserves the meaning of the original sentence.
For consistency, each pair is rated by three judges.
To ensure the quality of our evaluations, we selected workers who could reliably distinguish high from low agency sentences in a qualification task (see Figure \ref{fig:qual-task} in the appendix).

For this evaluation, we generate three revisions--one for each target agency level--for a random subset of 100 test examples.
We compare the output of our full \modelname{} model with two external baselines (\pplm and \bst). 
For further comparison, we also include the 
most competitive ablated baseline from Table~\ref{tab:results} (i.e., \joint+\noBoost).

\subsubsection{Results}
In Figure~\ref{fig:humaneval:H2H}, we show the percentages of times in which \modelname{} was preferred over the three baseline models.\footnote{Judgments in our evaluation task had an average pairwise agreement of 75\% (Krippendorf's $\alpha$=.52).}
Percentages $>$50\% indicate a preference towards \modelname.

Overall, the sentence revisions by \modelname are preferred over all of the baselines in obtaining the desired agency level.
For meaning preservation, our model is always selected over \bst, mirroring  BertScores in Table~\ref{tab:results:test}.
The difference is less stark when comparing to \pplm which sometimes makes no changes or irrelevant changes to the input sentence, and reversed when comparing to the ablated \noBoost.

Additionally, \bst revisions were marked as gibberish substantially more than those by other models ($63\%$ vs. 3-7$\%$).
While this seemingly contradicts \bst's low perplexity scores, this is in line with previous work showing automatic fluency metrics can favor degenerate, bland, or repetitive language \citep{nucleussampling}.



\section{Gender Bias in Movies}
\label{sec:movies}
As a proof-of-concept of \taskname, we 
investigate whether gender biases in portrayals of movie characters can be mitigated using \modelname.

\subsection{Movie Scripts Corpus}
We draw our data from the 767 modern English movie scripts by \citet{gorinski2015movie}, focusing on the narrations which describe characters and their actions (as opposed to the character's dialogue utterances).
Described in further detail in \S\ref{sup:movie-bias} in the appendix, we automatically extract characters and assign them a binary\footnote{Note that gender is a social construct that goes beyond the man-woman binary \cite{lorber1991social}, however more inclusive analyses (e.g., with non-binary genders) are not possible given the limited information about the individuals mentioned in our data.} gender (man, woman) using a list of highly gendered names (e.g., ``Sarah'', ``William'') and a list of gendered words (e.g., ``waiter,'' ``waitress'').
Following previous work \cite{Ramakrishna2017-cq,sap2017connotation}, we assign narration sentences to characters if their name appears in them.

Our corpus contains 16,763 characters from 767 different English movies.
Of those characters, 68\% are inferred to be men and only 32\% to be women,\footnote{There were 2597 characters for which the gender could not be inferred.} consistent with known gender skews in movie characters \cite{google2017genderbias}.
This bias in representation is also present at the narrative level.
Specifically, female characters are only mentioned in $n_{narr,f}=$27 narrations on average, compared to $n_{narr,m}=$34 narrations for male characters (Cohen's $|d|=0.13$, $p<0.001$).
Similarly, compared to their male counterparts, female characters are described in significantly fewer words ($n_{words,f}=329$, $n_{words,m}=435$, $|d|=0.14$, $p<0.001$) and with fewer verbs ($n_{verbs,f}=41$, $n_{verbs,m}=54$, $|d|=0.13$, $p<0.001$).

\subsection{Debiasing Portrayal in Movies}
\begin{figure}
    \centering
    \includegraphics[width=\columnwidth]{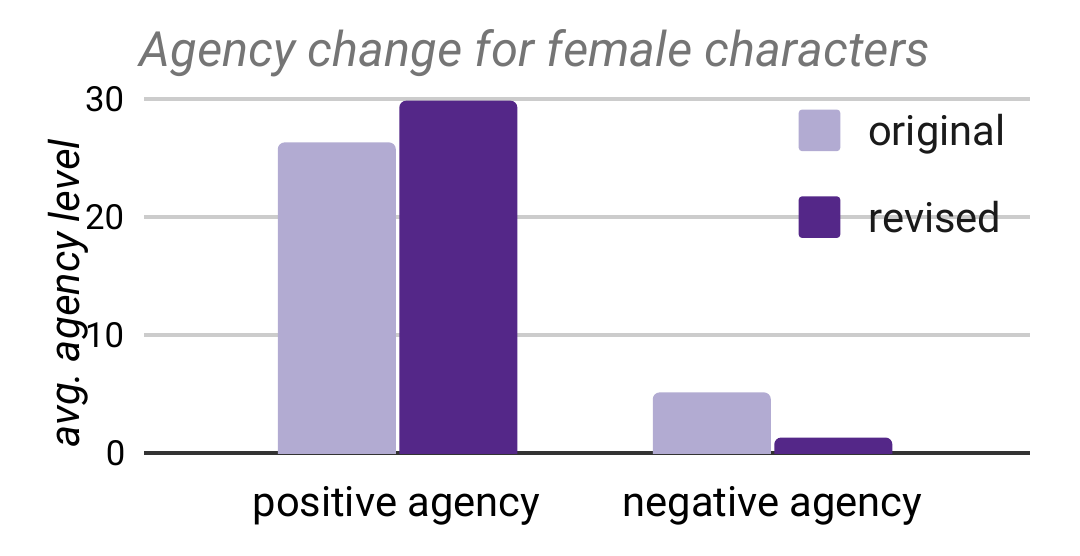}
    \caption{Average agency levels (i.e., number of agency verbs) for female characters in original and revised scripts. 
    \modelname can revise the portrayals of female characters in movies to give them higher positive agency and lower negative agency.}
    \label{fig:agency-women-movies}
\end{figure}

Given the known bias that female characters are portrayed with less agency \citep{sap2017connotation}, our goal is to re-balance their agency levels to be more on par with those of male characters.
Therefore, we revise only the sentences describing female characters to have higher agency, using \modelname.
Then we extract connotation frames of agency for revised script sentences, and aggregate per character.
Shown in Figure \ref{fig:agency-women-movies}, revisions successfully increase the instances of positive agency of female characters, and decrease their negative agency or passiveness.

We further examine the change in gender association of positive and negative agency, to verify the effectiveness of \taskname.
We first count all the positive and negative agency verbs used to describe characters (in original or rewritten sentences).
Following \citet{sap2017connotation}, we then fit a logistic regression model to quantify the association between character's gender with their agency levels, controlling for their number of words, verbs, and narrations.
For better interpretation of the $\beta$ coefficients, we $z$-score all the continuous variables.

We confirm that indeed, \taskname using \modelname can reverse the bias in portrayal in movies. 
In original scripts, male characters were portrayed with significantly higher positive agency ($\beta_{\text{pos}}=1.2$, $p<0.001$) and lower negative agency ($\beta_{\text{neg}}=-0.3$, $p<0.001$) than female characters.
However, our model successfully reverses this gender bias, portraying women with significantly more positive agency ($\beta'_{\text{pos}}=-62.6$, $p<0.001$) and significantly less negative agency ($\beta'_{\text{neg}}=8.7$, $p<0.001$).

Our findings on movie scripts show the promise of using \taskname to successfully mitigate gender biases in portrayal of characters, which could be extended to other domains \cite[e.g., news or fiction,][]{field-tsvetkov-2019-entity,fast2016shirtless}.
Additionally, future work could consider alternative views of portrayal biases \citep[e.g., ``regard'' or bias directed at different demographic groups;][]{sheng-etal-2019-woman,sap2020socialbiasframes}, or use more holistic views of gender roles \cite[e.g., ``masculine default'' cultures;][]{cheryan2020masculine}.

\section{Related Work}

\taskname is a new formalization of the unsupervised stylistic rewriting task, contrasting with supervised approaches which benefit from parallel corpora \cite[e.g., ][]{xu-etal-2012-paraphrasing,Xu-EtAl:2015:TACL,Rao2018-fd,pryzant2019automatically}.
In unsupervised settings, a majority of work has dealt with the dearth of parallel data by using encoder-decoder setups paired with discriminators to disentangle style from content and steer generations \cite[e.g.,][]{Shen2017-wf,Zhang2018-ke,Fu2018-el,Yang2018-gk,Niu2018-ag,Romanov2019-fb,Dai2019-uw,john-etal-2019-disentangled} or backtranslation setups \cite{Prabhumoye2018-kj,Lample2018-qa}.
In contrast, \citet{Li2018-op} introduce a modular approach \cite[later adapted to transformer models by][]{Sudhakar2019-fw} that relies on drop-in replacement of attribute markers followed by language correction.
\modelname improves on this approach with an additional out-of-domain paraphrasing objective.

While a majority of related existing stylistic rewriting work defines style as sentiment (e.g., on reviews), a notable exception is \citet{Nogueira_dos_Santos2018-yp}, who use stylistic rewriting to make text less hateful or offensive.
Similar in spirit, \taskname is a novel formalization that aims to address and revise social biases expressed in text, but using the nuanced implications distilled in connotation frames of power and agency instead of binary offensiveness.

Our work also draws inspiration from controllable generation methods \cite[e.g.,][]{koncel-kedziorski-etal-2016-theme,hu2017towardCG,Ficler2017-ab}. 
While those methods steer the generation output to contain desired attributes, controllable revision is constrained to revise an input sentence in addition to generating with desired attributes.









\section{Conclusion}

We introduce a new text revision task of \taskname, to help debias the portrayal of characters through the lens of connotation frames of power and agency.
To this end, we create \modelname, a transformer-based encoder-decoder trained on a joint reconstruction and paraphrasing objective.
Our approach demonstrates promising results to revise sentences with targeted power and agency, and outperforms ablations and baselines on both automatic and human evaluations.
Finally, as a case study, we show the feasibility for \taskname at debiasing the portrayal of characters in movie scripts.
Our findings highlight the potential of neural models as a tool for editing out social biases in text.

\section*{Acknowledgements}
The authors thank the anonymous reviewers and meta-reviewers for their helpful feedback.
We also thank Aishwarya Nirmal and Kenta Takatsu for their preliminary exploration of this task.
Additionally, we thank Ari Holtzman, Lucy Lin, Sofia Serrano, Elizabeth Clark, and other members of the UW NLP community for their thoughtful input.
This research was supported in part by NSF (IIS-1524371, IIS-1714566), DARPA under the CwC program through the ARO (W911NF-15-1-0543), DARPA under the MCS program through NIWC Pacific (N66001-19-2-4031), and the National Science Foundation Graduate Research Fellowship Program under Grant No. DGE-1256082.

\bibliography{09-references}

\begin{thebibliography}{57}
\expandafter\ifx\csname natexlab\endcsname\relax\def\natexlab#1{#1}\fi

\bibitem[{Bannard and Callison-Burch(2005)}]{bannard2005paraphrasing}
Colin Bannard and Chris Callison-Burch. 2005.
\newblock \href {https://www.aclweb.org/anthology/P05-1074} {Paraphrasing with
  bilingual parallel corpora}.
\newblock In \emph{ACL}.

\bibitem[{Behm-Morawitz and Mastro(2008)}]{behm2008mean}
Elizabeth Behm-Morawitz and Dana~E Mastro. 2008.
\newblock Mean girls? the influence of gender portrayals in teen movies on
  emerging adults' gender-based attitudes and beliefs.
\newblock \emph{Journalism \& Mass Communication Quarterly}, 85(1):131--146.

\bibitem[{Bird et~al.(2009)Bird, Klein, and Loper}]{bird2009natural}
Steven Bird, Ewan Klein, and Edward Loper. 2009.
\newblock \emph{Natural language processing with Python: analyzing text with
  the natural language toolkit}.
\newblock " O'Reilly Media, Inc.".

\bibitem[{Cao et~al.(2017)Cao, Luo, Li, and Li}]{seq2seqpara}
Ziqiang Cao, Chuwei Luo, Wenjie Li, and Sujian Li. 2017.
\newblock \href {https://aaai.org/ocs/index.php/AAAI/AAAI17/paper/view/14527}
  {Joint copying and restricted generation for paraphrase}.
\newblock In \emph{AAAI}.

\bibitem[{Cheryan and Markus(2020)}]{cheryan2020masculine}
Sapna Cheryan and Hazel~Rose Markus. 2020.
\newblock Masculine defaults: Identifying and mitigating hidden cultural
  biases.
\newblock \emph{Psychological Review}.

\bibitem[{Clark et~al.(2018)Clark, Ross, Tan, Ji, and
  Smith}]{clark2018creative}
Elizabeth Clark, Anne~Spencer Ross, Chenhao Tan, Yangfeng Ji, and Noah~A Smith.
  2018.
\newblock \href {https://dl.acm.org/doi/10.1145/3172944.3172983} {Creative
  writing with a machine in the loop: Case studies on slogans and stories}.
\newblock In \emph{IUI}.

\bibitem[{Creutz(2018)}]{Creutz2018opuparcus}
Mathias Creutz. 2018.
\newblock \href {https://www.aclweb.org/anthology/L18-1218} {Open subtitles
  paraphrase corpus for six languages}.
\newblock In \emph{{LREC}}.
\newblock Corpus available at \url{http://urn.fi/urn:nbn:fi:lb-201804191}.

\bibitem[{Dai et~al.(2019)Dai, Liang, Qiu, and Huang}]{Dai2019-uw}
Ning Dai, Jianze Liang, Xipeng Qiu, and Xuanjing Huang. 2019.
\newblock \href {https://www.aclweb.org/anthology/P19-1601} {Style transformer:
  Unpaired text style transfer without disentangled latent representation}.
\newblock In \emph{{ACL}}.

\bibitem[{Dathathri et~al.(2020)Dathathri, Madotto, Lan, Hung, Frank, Molino,
  Yosinski, and Liu}]{Dathathri2020-ua}
Sumanth Dathathri, Andrea Madotto, Janice Lan, Jane Hung, Eric Frank, Piero
  Molino, Jason Yosinski, and Rosanne Liu. 2020.
\newblock \href {https://openreview.net/forum?id=H1edEyBKDS} {Plug and play
  language models: A simple approach to controlled text generation}.
\newblock In \emph{{ICLR}}.

\bibitem[{Dennett(1989)}]{dennett1989intentional}
Daniel~Clement Dennett. 1989.
\newblock \emph{The intentional stance}.
\newblock MIT press.

\bibitem[{Fast et~al.(2016)Fast, Vachovsky, and Bernstein}]{fast2016shirtless}
Ethan Fast, Tina Vachovsky, and Michael~S Bernstein. 2016.
\newblock \href {https://arxiv.org/abs/1603.08832} {Shirtless and dangerous:
  Quantifying linguistic signals of gender bias in an online fiction writing
  community}.
\newblock In \emph{{ICWSM}}.

\bibitem[{Ficler and Goldberg(2017)}]{Ficler2017-ab}
Jessica Ficler and Yoav Goldberg. 2017.
\newblock \href {https://www.aclweb.org/anthology/W17-4912} {Controlling
  linguistic style aspects in neural language generation}.
\newblock In \emph{{EMNLP} Workshop on Stylistic Variation}.

\bibitem[{Field et~al.(2019)Field, Bhat, and Tsvetkov}]{field2019contextual}
Anjalie Field, Gayatri Bhat, and Yulia Tsvetkov. 2019.
\newblock \href {https://www.aaai.org/ojs/index.php/ICWSM/article/view/3358}
  {Contextual affective analysis: A case study of people portrayals in online
  \#metoo stories}.
\newblock In \emph{ICWSM}.

\bibitem[{Field and Tsvetkov(2019)}]{field-tsvetkov-2019-entity}
Anjalie Field and Yulia Tsvetkov. 2019.
\newblock \href {https://www.aclweb.org/anthology/P19-1243} {Entity-centric
  contextual affective analysis}.
\newblock In \emph{ACL}.

\bibitem[{Fiske(1993)}]{Fiske1993controlling}
Susan~T Fiske. 1993.
\newblock Controlling other people. the impact of power on stereotyping.
\newblock \emph{American psychologist}, 48(6):621--628.

\bibitem[{Fu et~al.(2018)Fu, Tan, Peng, Zhao, and Yan}]{Fu2018-el}
Zhenxin Fu, Xiaoye Tan, Nanyun Peng, Dongyan Zhao, and Rui Yan. 2018.
\newblock \href
  {https://www.aaai.org/ocs/index.php/AAAI/AAAI18/paper/view/17015} {Style
  transfer in text: Exploration and evaluation}.
\newblock In \emph{{AAAI}}.

\bibitem[{Ghazvininejad et~al.(2016)Ghazvininejad, Shi, Choi, and
  Knight}]{ghazvininejad2016generating}
Marjan Ghazvininejad, Xing Shi, Yejin Choi, and Kevin Knight. 2016.
\newblock \href {https://www.aclweb.org/anthology/D16-1126} {Generating topical
  poetry}.
\newblock In \emph{EMNLP}.

\bibitem[{Ghazvininejad et~al.(2017)Ghazvininejad, Shi, Priyadarshi, and
  Knight}]{ghazvininejad2017hafez}
Marjan Ghazvininejad, Xing Shi, Jay Priyadarshi, and Kevin Knight. 2017.
\newblock \href {https://www.aclweb.org/anthology/P17-4008} {{H}afez: an
  interactive poetry generation system}.
\newblock In \emph{ACL Demonstrations}.

\bibitem[{Ghosh et~al.(2017)Ghosh, Chollet, Laksana, Morency, and
  Scherer}]{ghosh2017affectlm}
Sayan Ghosh, Mathieu Chollet, Eugene Laksana, Louis-Philippe Morency, and
  Stefan Scherer. 2017.
\newblock \href {https://www.aclweb.org/anthology/P17-1059} {{Affect-LM}: A
  neural language model for customizable affective text generation}.
\newblock In \emph{{ACL}}.

\bibitem[{{Google}(2017)}]{google2017genderbias}
{Google}. 2017.
\newblock Using technology to address gender bias in film.
\newblock
  \url{https://www.google.com/about/main/gender-equality-films/index.html}.

\bibitem[{Gorinski and Lapata(2015)}]{gorinski2015movie}
Philip Gorinski and Mirella Lapata. 2015.
\newblock \href {https://www.aclweb.org/anthology/N15-1113} {Movie script
  summarization as graph-based scene extraction}.
\newblock In \emph{NAACL}.

\bibitem[{Holtzman et~al.(2020)Holtzman, Buys, Du, Forbes, and
  Choi}]{nucleussampling}
Ari Holtzman, Jan Buys, Li~Du, Maxwell Forbes, and Yejin Choi. 2020.
\newblock \href {https://openreview.net/forum?id=rygGQyrFvH} {The curious case
  of neural text degeneration}.
\newblock In \emph{ICLR}.

\bibitem[{Hu et~al.(2017)Hu, Yang, Liang, Salakhutdinov, and
  Xing}]{hu2017towardCG}
Zhiting Hu, Zichao Yang, Xiaodan Liang, Ruslan Salakhutdinov, and Eric~P Xing.
  2017.
\newblock \href {http://proceedings.mlr.press/v70/hu17e.html} {Toward
  controlled generation of text}.
\newblock In \emph{ICML}.

\bibitem[{John et~al.(2019)John, Mou, Bahuleyan, and
  Vechtomova}]{john-etal-2019-disentangled}
Vineet John, Lili Mou, Hareesh Bahuleyan, and Olga Vechtomova. 2019.
\newblock \href {https://doi.org/10.18653/v1/P19-1041} {Disentangled
  representation learning for non-parallel text style transfer}.
\newblock In \emph{ACL}.

\bibitem[{Kiritchenko and Mohammad(2017)}]{kiritchenko-mohammad-2017-best}
Svetlana Kiritchenko and Saif Mohammad. 2017.
\newblock \href {https://www.aclweb.org/anthology/P17-2074} {Best-worst scaling
  more reliable than rating scales: A case study on sentiment intensity
  annotation}.
\newblock In \emph{ACL}.

\bibitem[{Koncel-Kedziorski et~al.(2016)Koncel-Kedziorski, Konstas,
  Zettlemoyer, and Hajishirzi}]{koncel-kedziorski-etal-2016-theme}
Rik Koncel-Kedziorski, Ioannis Konstas, Luke Zettlemoyer, and Hannaneh
  Hajishirzi. 2016.
\newblock \href {https://doi.org/10.18653/v1/D16-1168} {A theme-rewriting
  approach for generating algebra word problems}.
\newblock In \emph{EMNLP}.

\bibitem[{Lakoff(1973)}]{lakoff1973language}
Robin Lakoff. 1973.
\newblock Language and woman's place.
\newblock \emph{Language in society}, 2(1):45--79.

\bibitem[{Lample et~al.(2018)Lample, Subramanian, Smith, Denoyer, Ranzato, and
  Boureau}]{Lample2018-qa}
Guillaume Lample, Sandeep Subramanian, Eric Smith, Ludovic Denoyer,
  Marc'aurelio Ranzato, and Y-Lan Boureau. 2018.
\newblock \href {https://openreview.net/forum?id=H1g2NhC5KQ}
  {{Multiple-Attribute} text rewriting}.
\newblock In \emph{{ICLR}}.

\bibitem[{Li et~al.(2018{\natexlab{a}})Li, Jia, He, and Liang}]{Li2018-op}
Juncen Li, Robin Jia, He~He, and Percy Liang. 2018{\natexlab{a}}.
\newblock \href {https://www.aclweb.org/anthology/N18-1169} {Delete, retrieve,
  generate: A simple approach to sentiment and style transfer}.
\newblock In \emph{{NAACL}}.

\bibitem[{Li et~al.(2018{\natexlab{b}})Li, Jiang, Shang, and
  Li}]{li2018paraphrase}
Zichao Li, Xin Jiang, Lifeng Shang, and Hang Li. 2018{\natexlab{b}}.
\newblock \href {https://www.aclweb.org/anthology/D18-1421/} {Paraphrase
  generation with deep reinforcement learning}.
\newblock In \emph{EMNLP}.

\bibitem[{Liu et~al.(2016)Liu, Lowe, Serban, Noseworthy, Charlin, and
  Pineau}]{HowNotToEval}
Chia-Wei Liu, Ryan Lowe, Iulian Serban, Mike Noseworthy, Laurent Charlin, and
  Joelle Pineau. 2016.
\newblock \href {https://www.aclweb.org/anthology/D16-1230} {How {NOT} to
  evaluate your dialogue system: An empirical study of unsupervised evaluation
  metrics for dialogue response generation}.
\newblock In \emph{EMNLP}.

\bibitem[{Lorber et~al.(1991)Lorber, Farrell et~al.}]{lorber1991social}
Judith Lorber, Susan~A Farrell, et~al. 1991.
\newblock The social construction of gender.
\newblock \emph{Newbury Park}, 5.

\bibitem[{Loshchilov and Hutter(2019)}]{Loshchilov2019DecoupledWD}
Ilya Loshchilov and Frank Hutter. 2019.
\newblock \href {https://openreview.net/forum?id=Bkg6RiCqY7} {Decoupled weight
  decay regularization}.
\newblock In \emph{ICLR}.

\bibitem[{Mir et~al.(2019)Mir, Felbo, Obradovich, and Rahwan}]{Mir2019-bv}
Remi Mir, Bjarke Felbo, Nick Obradovich, and Iyad Rahwan. 2019.
\newblock \href {https://www.aclweb.org/anthology/N19-1049} {Evaluating style
  transfer for text}.
\newblock In \emph{{NAACL}}.

\bibitem[{Mostafazadeh et~al.(2016)Mostafazadeh, Chambers, He, Parikh, Batra,
  Vanderwende, Kohli, and Allen}]{Mostafazadeh2016ROCStory}
Nasrin Mostafazadeh, Nathanael Chambers, Xiaodong He, Devi Parikh, Dhruv Batra,
  Lucy Vanderwende, Pushmeet Kohli, and James Allen. 2016.
\newblock \href {https://www.aclweb.org/anthology/N16-1098} {A corpus and cloze
  evaluation for deeper understanding of commonsense stories}.
\newblock In \emph{{NAACL}}.
\newblock Corpus available at
  \url{https://www.cs.rochester.edu/nlp/rocstories/}.

\bibitem[{Niu and Bansal(2018)}]{Niu2018-ag}
Tong Niu and Mohit Bansal. 2018.
\newblock \href {https://www.aclweb.org/anthology/Q18-1027} {Polite dialogue
  generation without parallel data}.
\newblock \emph{TACL}.

\bibitem[{Pennington et~al.(2014)Pennington, Socher, and Manning}]{glove}
Jeffrey Pennington, Richard Socher, and Christopher~D. Manning. 2014.
\newblock \href {http://www.aclweb.org/anthology/D14-1162} {Glove: Global
  vectors for word representation}.
\newblock In \emph{EMNLP}.

\bibitem[{Prabhumoye et~al.(2018)Prabhumoye, Tsvetkov, Salakhutdinov, and
  Black}]{Prabhumoye2018-kj}
Shrimai Prabhumoye, Yulia Tsvetkov, Ruslan Salakhutdinov, and Alan~W Black.
  2018.
\newblock \href {https://www.aclweb.org/anthology/P18-1080} {Style transfer
  through {Back-Translation}}.
\newblock In \emph{{ACL}}.
\newblock Code available at
  \url{https://github.com/shrimai/Style-Transfer-Through-Back-Translation}.

\bibitem[{Prakash et~al.(2016)Prakash, Hasan, Lee, Datla, Qadir, Liu, and
  Farri}]{prakash-paraphrase}
Aaditya Prakash, Sadid~A. Hasan, Kathy Lee, Vivek Datla, Ashequl Qadir, Joey
  Liu, and Oladimeji Farri. 2016.
\newblock \href {https://www.aclweb.org/anthology/C16-1275} {Neural paraphrase
  generation with stacked residual {LSTM} networks}.
\newblock In \emph{COLING}.

\bibitem[{Pryzant et~al.(2020)Pryzant, Martinez, Dass, Kurohashi, Jurafsky, and
  Yang}]{pryzant2019automatically}
Reid Pryzant, Richard~Diehl Martinez, Nathan Dass, Sadao Kurohashi, Dan
  Jurafsky, and Diyi Yang. 2020.
\newblock \href {https://aaai.org/ojs/index.php/AAAI/article/view/5385}
  {Automatically neutralizing subjective bias in text}.
\newblock In \emph{AAAI}.

\bibitem[{Radford et~al.(2018)Radford, Narasimhan, Salimans, and
  Sutskever}]{radford2018improving}
Alec Radford, Karthik Narasimhan, Tim Salimans, and Ilya Sutskever. 2018.
\newblock \href
  {https://s3-us-west-2.amazonaws.com/openai-assets/researchcovers/languageunsupervised/language
  understanding paper.pdf} {Improving language understanding by generative
  pre-training}.
\newblock Unpublished.

\bibitem[{Ramakrishna et~al.(2017)Ramakrishna, Mart{\'\i}nez, Malandrakis,
  Singla, and Narayanan}]{Ramakrishna2017-cq}
Anil Ramakrishna, Victor~R Mart{\'\i}nez, Nikolaos Malandrakis, Karan Singla,
  and Shrikanth Narayanan. 2017.
\newblock \href {https://www.aclweb.org/anthology/P17-1153} {Linguistic
  analysis of differences in portrayal of movie characters}.
\newblock In \emph{{ACL}}.

\bibitem[{Rao and Tetreault(2018)}]{Rao2018-fd}
Sudha Rao and Joel Tetreault. 2018.
\newblock \href {https://www.aclweb.org/anthology/N18-1012} {Dear sir or madam,
  may {I} introduce the {GYAFC} dataset: Corpus, benchmarks and metrics for
  formality style transfer}.
\newblock In \emph{{NAACL}}.

\bibitem[{Rashkin et~al.(2016)Rashkin, Singh, and
  Choi}]{rashkin2016connotationframes}
Hannah Rashkin, Sameer Singh, and Yejin Choi. 2016.
\newblock \href {https://doi.org/10.18653/v1/P16-1030} {Connotation frames: A
  data-driven investigation}.
\newblock In \emph{ACL}.

\bibitem[{Romanov et~al.(2019)Romanov, Rumshisky, Rogers, and
  Donahue}]{Romanov2019-fb}
Alexey Romanov, Anna Rumshisky, Anna Rogers, and David Donahue. 2019.
\newblock \href {https://www.aclweb.org/anthology/N19-1088} {Adversarial
  decomposition of text representation}.
\newblock In \emph{{NAACL}}.

\bibitem[{Nogueira~dos Santos et~al.(2018)Nogueira~dos Santos, Melnyk, and
  Padhi}]{Nogueira_dos_Santos2018-yp}
Cicero Nogueira~dos Santos, Igor Melnyk, and Inkit Padhi. 2018.
\newblock \href {https://www.aclweb.org/anthology/P18-2031/} {Fighting
  offensive language on social media with unsupervised text style transfer}.
\newblock In \emph{{ACL}}.

\bibitem[{Sap et~al.(2020)Sap, Gabriel, Qin, Jurafsky, Smith, and
  Choi}]{sap2020socialbiasframes}
Maarten Sap, Saadia Gabriel, Lianhui Qin, Dan Jurafsky, Noah~A Smith, and Yejin
  Choi. 2020.
\newblock \href {https://www.aclweb.org/anthology/2020.acl-main.486} {Social
  bias frames: Reasoning about social and power implications of language}.
\newblock In \emph{ACL}.

\bibitem[{Sap et~al.(2017)Sap, Prasettio, Holtzman, Rashkin, and
  Choi}]{sap2017connotation}
Maarten Sap, Marcella~Cindy Prasettio, Ari Holtzman, Hannah Rashkin, and Yejin
  Choi. 2017.
\newblock \href {https://www.aclweb.org/anthology/D17-1247} {Connotation frames
  of power and agency in modern films}.
\newblock In \emph{EMNLP}.
\newblock Connotation Frames downloaded from
  \url{http://maartensap.com/movie-bias/}.

\bibitem[{Shen et~al.(2017)Shen, Lei, Barzilay, and Jaakkola}]{Shen2017-wf}
Tianxiao Shen, Tao Lei, Regina Barzilay, and Tommi Jaakkola. 2017.
\newblock \href
  {http://papers.neurips.cc/paper/7259-style-transfer-from-non-parallel-text-by-cross-alignment}
  {Style transfer from {Non-Parallel} text by {Cross-Alignment}}.
\newblock In \emph{{NeurIPS}}.

\bibitem[{Sheng et~al.(2019)Sheng, Chang, Natarajan, and
  Peng}]{sheng-etal-2019-woman}
Emily Sheng, Kai-Wei Chang, Premkumar Natarajan, and Nanyun Peng. 2019.
\newblock \href {https://doi.org/10.18653/v1/D19-1339} {The woman worked as a
  babysitter: On biases in language generation}.
\newblock In \emph{EMNLP}.

\bibitem[{Sudhakar et~al.(2019)Sudhakar, Upadhyay, and
  Maheswaran}]{Sudhakar2019-fw}
Akhilesh Sudhakar, Bhargav Upadhyay, and Arjun Maheswaran. 2019.
\newblock \href {https://www.aclweb.org/anthology/D19-1322} {Transforming
  delete, retrieve, generate approach for controlled text style transfer}.
\newblock In \emph{{EMNLP}}.

\bibitem[{Wolf et~al.(2019)Wolf, Debut, Sanh, Chaumond, Delangue, Moi, Cistac,
  Rault, Louf, Funtowicz et~al.}]{wolf2019huggingface}
Thomas Wolf, L~Debut, V~Sanh, J~Chaumond, C~Delangue, A~Moi, P~Cistac, T~Rault,
  R~Louf, M~Funtowicz, et~al. 2019.
\newblock \href {https://arxiv.org/abs/1910.03771} {Huggingface’s
  transformers: State-of-the-art natural language processing}.
\newblock Unpublished.

\bibitem[{Xu et~al.(2015)Xu, Callison-Burch, and Napoles}]{Xu-EtAl:2015:TACL}
Wei Xu, Chris Callison-Burch, and Courtney Napoles. 2015.
\newblock \href {https://www.aclweb.org/anthology/Q15-1021} {Problems in
  current text simplification research: New data can help}.
\newblock \emph{TACL}.

\bibitem[{Xu et~al.(2012)Xu, Ritter, Dolan, Grishman, and
  Cherry}]{xu-etal-2012-paraphrasing}
Wei Xu, Alan Ritter, Bill Dolan, Ralph Grishman, and Colin Cherry. 2012.
\newblock \href {https://www.aclweb.org/anthology/C12-1177} {Paraphrasing for
  style}.
\newblock In \emph{COLING}.

\bibitem[{Yang et~al.(2018)Yang, Hu, Dyer, Xing, and
  Berg-Kirkpatrick}]{Yang2018-gk}
Zichao Yang, Zhiting Hu, Chris Dyer, Eric~P Xing, and Taylor Berg-Kirkpatrick.
  2018.
\newblock \href
  {https://papers.neurips.cc/paper/7959-unsupervised-text-style-transfer-using-language-models-as-discriminators}
  {Unsupervised text style transfer using language models as discriminators}.
\newblock In \emph{{NeurIPS}}.

\bibitem[{Zhang et~al.(2020)Zhang, Kishore, Wu, Weinberger, and
  Artzi}]{bertscoreppr}
Tianyi Zhang, Varsha Kishore, Felix Wu, Kilian~Q. Weinberger, and Yoav Artzi.
  2020.
\newblock \href {https://openreview.net/forum?id=SkeHuCVFDr} {Bertscore:
  Evaluating text generation with {BERT}}.
\newblock In \emph{ICLR}.

\bibitem[{Zhang et~al.(2018)Zhang, Ding, and Soricut}]{Zhang2018-ke}
Ye~Zhang, Nan Ding, and Radu Soricut. 2018.
\newblock \href {https://www.aclweb.org/anthology/N18-1138} {{SHAPED}:
  {Shared-Private} {Encoder-Decoder} for text style adaptation}.
\newblock In \emph{{NAACL}}.

\end{thebibliography}
\bibliographystyle{acl_natbib}

\clearpage
\appendix
\section{Additional data description}
\label{sup:data-deets}
\subsection{ROC story corpus}
This English corpus originally contains 100,000 five-sentence stories written by crowdworkers about realistic everyday scenarios. 
We select the data for our task by first extracting agency levels for each sentence, filtering out those with indeterminable agency.
Additionally, we filter out sentences with four or more verbs, to prevent the sentence masking from deleting too many content words.

\label{sup:roc-deets}
\subsection{Paraphrase corpus}
\label{sup:para-deets}
This corpus contains paraphrases of spoken dialogue extracted from movie and TV subtitles.\footnote{From \url{http://www.opensubtitles.org}}
OpusParcus was created by automatically aligning the subtitles sentences using several probabilistic metrics, including likelihood under a round-trip translation paraphrasing model \cite{bannard2005paraphrasing} and pointwise mutual information.
For our paraphrasing dataset, we apply the same filtering as with the ROC story corpus to the English portion of the OpusParcus training corpus and select the top 10\% highest scoring paraphrases using the PMI scoring from the original paper.
We extract agency levels for each pair of paraphrases, and select pairs to obtain roughly equal number of agency-level pairs (i.e., 1/9th positive-neutral, 1/9th positive-negative, etc.)
We preprocess the text by stripping any leading periods and commas.

\section{Experimental details}
\label{sup:experimental-details}
We use the Hugging Face \cite{wolf2019huggingface} implementation of OpenAI's GPT model \cite[117M parameters;][]{radford2018improving}.
our final setup uses AdamW \cite{Loshchilov2019DecoupledWD} as our optimizer with a learning weight of 1e-5, batch size of 4 and maximum sequence length of 64.
In preliminary results, we find that $\beta$=5 aptly steers the generation while avoiding repetition issues.

\subsection{\modelname details}
\label{sup:model-hyperparams}

\begin{table}[t]
    \centering
    \begin{tabular}{cc}
    \toprule
        \textbf{Hyperparameter} & \textbf{Value} \\ \midrule
        Vocabulary Size & 40486\\
        Maximum Sequence Length & 64 \\
        Training Batch Size & 4\\
        Embedding Size & 768\\
        \# Attention Heads & 12\\
        \# Attention Layers & 12\\
        \bottomrule
    \end{tabular}
         
     
    \caption{\modelname hyperparameters.}
    \label{tab:hyperparams}
\end{table}

All the experiments are performed on NVIDIA TITAN card and use the model hyperparameters listed in Table \ref{tab:hyperparams}.

\subsubsection{\modelshort$_{ParaOnly+None}$} 
We train this model for 10 epochs with each epoch taking approximately an hour. The learning rate is 1e-5 with AdamW optimizer, which is tuned manually in the [1e-6, 1e-3] range for 7 times. We use $p=0.4$ for nucleus sampling and $p$ is tuned manually in the [0.4, 0.9] range for 5 values.

\subsubsection{\modelshort$_{ParaOnly+Static}$}
The \modelshort$_{ParaOnly+Static}$ loads the trained model from \modelshort$_{ParaOnly+None}$ and add re-scaling to the logits. The re-scaling factor, $\beta$ was tuned manually tuned in the [0, 10] range. We try 8 $\beta$s and use 5 in the final model. We use the same $p$ as \modelshort$_{ParaOnly+None}$

\subsubsection{\modelshort$_{Joint+None}$}
Similar to \modelshort$_{ParaOnly+None}$, we train this model for 10 epochs with each epoch taking approximately an hour. The learning rate is 1e-5 with AdamW optimizer, which is tuned manually in the [1e-6, 1e-3] range for 7 times.We use the same $p$ as \modelshort$_{ParaOnly+None}$

\subsubsection{\modelshort$_{Joint+Static}$}
The \modelshort$_{Joint+Static}$ loads the trained model from \modelshort$_{Joint+None}$ and add re-scaling to the logits. The re-scaling factor, $\beta$ was tuned manually tuned in the [0, 10] range. We try 8 $\beta$s and use 5 in the final model. We use the same $p$ as \modelshort$_{ParaOnly+None}$

\subsection{PPLM details}
\label{sup:pplm-details}
The PPLM decoding method can be used on top of any model, but their original codebase is for use with a pre-trained language model rather than a model for paraphrasing or style transfer.  We augment their techniques for this task by replacing the base model in their code with a denoising autoencoder that was trained to reconstruct the input sentence.   The denoising autoencoder was implemented using the base GPT2 model (to fit with their code library and be similar size to our model).
It was trained on our ROC only training data with a reconstruction objective.
In order to denoise the autoencoder, we randomly ``dropout'' about 50\% of the tokens from the context by replacing them with mask tokens.  This autoencoder is trained to reconstruct input sentences, but when used with the PPLM decoding method, the input gets dynamically updated to decode a sentence that is similar in meaning but more likely to have a positive/negative agency according to a discriminator that is trained on top of the autoencoder. The PPLM decoding method also has  hyperparameters that control the strength of the target label. If set too high, then the output could be degenerate.  We manually set the hyperparameters to be as strong possible without producing degenerate text, using a subset of the dev. set as a guide.

\subsection{Backtranslation details}
\label{sup:bst-details}
We use the code provided by \citet{Prabhumoye2018-kj} for running this baseline.
After lowercasing all the negative and positive agency examples in our training data (ROC and OpusParcus), we translate to French using the machine translation model provided in the code base.
This baseline requires training a style classifier (agency) and two decoders (one for each agency level). 
Since the classifier essentially re-learns the agency lexicon, we do not search for hyperparameters, and simply set a learning rate of 5, and 6 epochs.
For training the decoders, we perform grid search to find the best hyperparameters.
We experiment with a learning rates of \{0.5, 1, 2, 5\}, \{2, 3, 5\} epochs, a classification-loss weight of \{0.5, 1, 2\}, and a word-loss weight of \{0.5, 1, 2\}, and select the configuration with the best word-level accuracy on the dev. set.
We use SGD with a batch size of 64 for all experiments, and refer the reader to the code base for other default parameters.

\begin{figure*}
    \centering
    \fbox{\includegraphics[width=.85\textwidth]{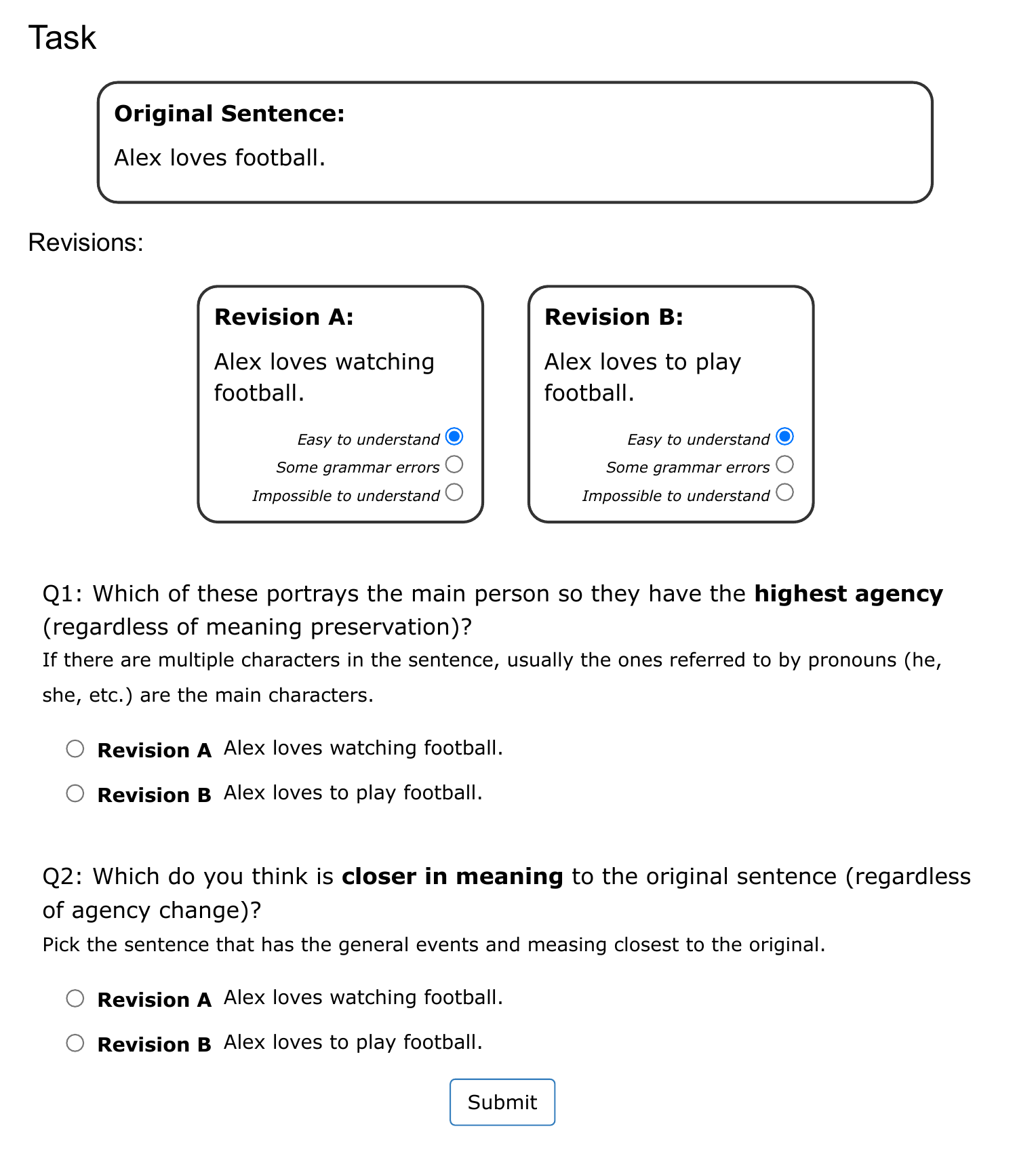}}
    \caption{Screenshot of the human evaluation annotation task.}
    \label{fig:mturk-task}
\end{figure*}

\begin{figure*}
    \centering
    \fbox{\includegraphics[width=.85\textwidth]{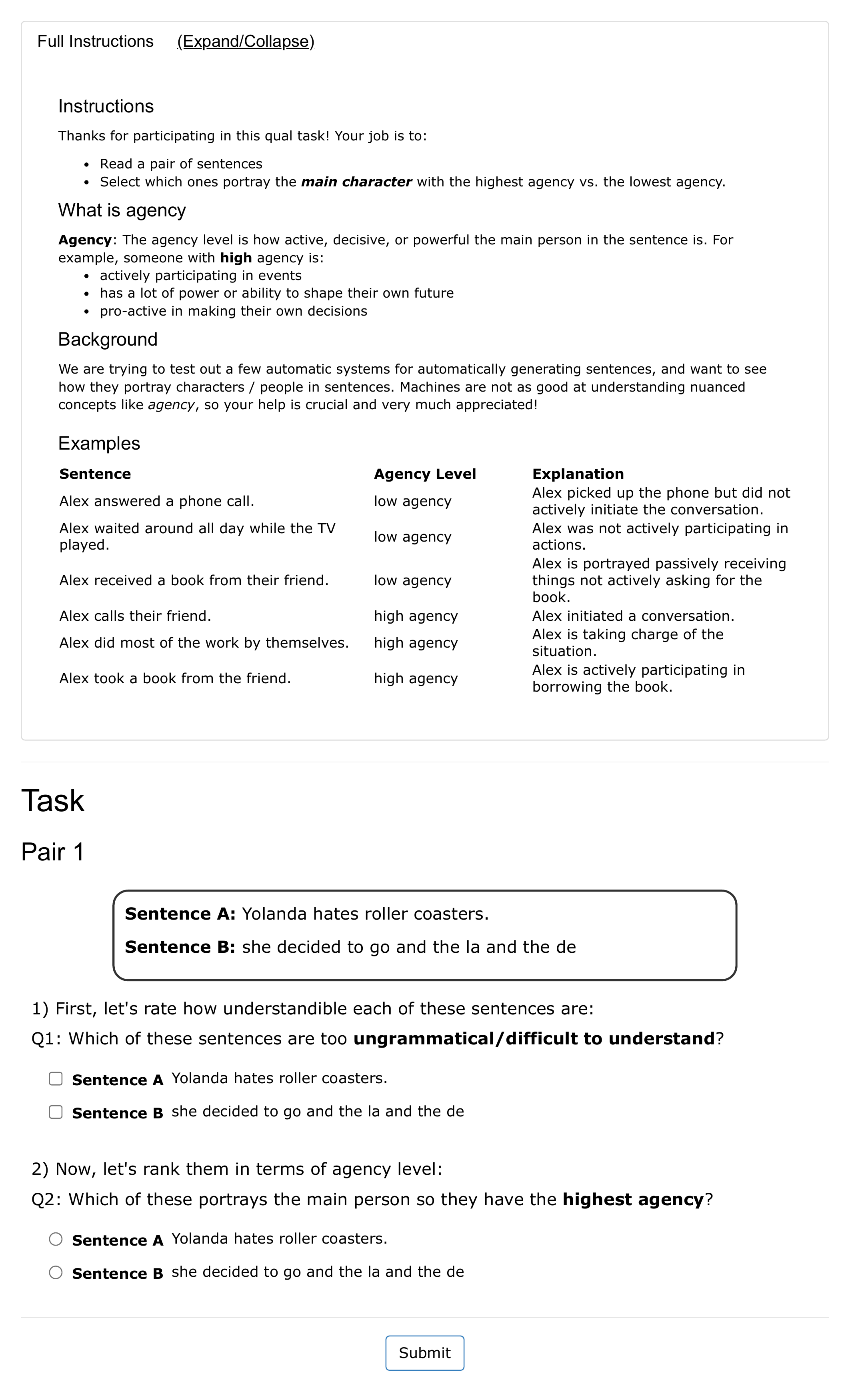}}
    \caption{Screenshot of the qualification task and its instructions. In the real task, workers rated three pairs of sentences, but only one is shown here.}
    \label{fig:qual-task}
\end{figure*}

\begin{table*}
    \centering
    \small
    \begin{tabular}{@{}c@{\hspace{.2em}}c@{\hspace{.5em}}p{2.25cm}@{\hspace{.7em}}r@{\hspace{.7em}}p{6cm}@{}@{\hspace{.1em}}c@{}}
    \toprule
          &$\Delta$Agency & Input & Model & Output & Agency(out)\\    
        \midrule
        \multirow{8}{*}{(a)} & \multirow{8}{*}{\high{+}~$\rightarrow$~~\low{--}} & \multirow{8}{2.5cm}{After the party I headed home.}
        & PPLM & After the party my classmate, Kayla and the Tgirls of the Universe. & =\\
        && & BST & please 's , i have a word of this . & \low{--}\\
        &&& \modelshort$_{ParaOnly+NoBoost}$ & after the party i headed home. & \high{+}\\
        &&& \modelshort$_{ParaOnly+Boost}$ & after the party i headed home. & \high{+}\\
        &&& \modelshort$_{Joint+SupplyVerb}$ & after the party i faced home. & \low{--}\\
        & & & \modelshort$_{Joint+NoBoost}$ & after the party i stayed home. & \low{--}\\
        & & & \modelshort$_{Joint+Boost}$ & after the party i stayed home. & \low{--}\\
         \midrule
         \multirow{13}{*}{(b)}&\multirow{13}{*}{ \high{+}~$\rightarrow$~~\low{--} } & 
         \multirow{13}{2.25cm}{A friend asked me to watch her two year old child for a minute.} 
        & PPLM & A Friend asked me to watch her two year old child for a minute. & \high{+}\\
        & & & BST & l didn 't have a word of this , you 're . & \low{--}\\ 
        &&& \modelshort$_{ParaOnly+NoBoost}$ & a friend asked me to watch her two year old child for a minute. & \high{+}\\
        &&& \modelshort$_{ParaOnly+Boost}$ & a friend asked me to watch her two year old child for a minute. & \high{+}\\
        &&& \modelshort$_{Joint+SupplyVerb}$ & a friend told me to watch her two year old child for a minute. & \high{+}\\
        & & & \modelshort$_{Joint+NoBoost}$ &a friend needed me to watch her two year old child for a minute. &\low{--}\\
        & & & \modelshort$_{Joint+Boost}$ & a friend {needed} me to watch her two year old child for a minute. &\low{--}\\
         \midrule
         \multirow{8}{*}{(c)} & \multirow{8}{*}{\high{+}~$\rightarrow$~~\low{--}} &
         \multirow{8}{2.25cm}{After filling in the data it looked quite sharp.}
         & PPLM & Before filling the last question it it it it looked quite sharp. Before filling the last question it it & \high{+}\\
         & & & BST & when the 't you want a word ? & \low{--}\\
         &&& \modelshort$_{ParaOnly+NoBoost}$ & after filling in the data it looked quite sharp. & \high{+}\\
         &&& \modelshort$_{ParaOnly+Boost}$ & after filling in the data it seemed quite sharp. & \high{+}\\
         &&& \modelshort$_{Joint+SupplyVerb}$ & after putting in the data it looked quite sharp. & = \\
         & & & \modelshort$_{Joint+NoBoost}$ & after analyzing in the data it looked quite sharp. & =\\
         & & & \modelshort$_{Joint+Boost}$ & after seeing in the data it seemed quite sharp. & \low{--}\\
          \midrule
         \multirow{7}{*}{(d)} & \multirow{7}{*}{\low{--}~~$\rightarrow$~\high{+}} &
         \multirow{7}{2.25cm}{Allie was failing science class.}
         & PPLM & Allie was failing science grade. & \low{--}\\
         & & & BST & do you want me ? & \high{+}\\
         &&& \modelshort$_{ParaOnly+NoBoost}$ & allie was failing science class.& \low{--} \\
         &&& \modelshort$_{ParaOnly+Boost}$ & allie was failing science class.& \low{--} \\
         &&& \modelshort$_{Joint+SupplyVerb}$ & allie was ignoring science class. & \high{+}\\
         & & & \modelshort$_{Joint+NoBoost}$ & allie was failing science class. & \low{--}\\
         & & & \modelshort$_{Joint+Boost}$ & allie was taking science class. & \high{+}\\
         \midrule
        \multirow{7}{*}{(e)} & \multirow{7}{*}{\low{--}~~$\rightarrow$~\high{+}}&         \multirow{7}{2.25cm}{Darla wanted a soft drink.}
         & PPLM & darla wants a hard hard drink. & \low{--}\\
        & & & BST & don 't take me a man . & \high{+}\\
        & & & \modelshort$_{ParaOnly+NoBoost}$ & darla wanted a soft drink. & \low{--}\\
        & & & \modelshort$_{ParaOnly+Boost}$ & darla wanted a soft drink. & \low{--}\\
        & & & \modelshort$_{Joint+SupplyVerb}$ & darla got a soft drink. & \high{+}\\
        & & & \modelshort$_{Joint+NoBoost}$ & darla ordered a soft drink. & \high{+}\\
        & & & \modelshort$_{Joint+Boost}$ & darla ordered a soft drink. & \high{+}\\
          \midrule
          
         \multirow{7}{*}{(f)}&\multirow{7}{*}{ \low{--}~~$\rightarrow$~\high{+} }& \multirow{7}{2.25cm}{Clint paused on the trail.} 
          & PPLM & clint was on the trail. & \\
            & & & BST & don 't you want me , & \low{--}\\
            & & & \modelshort$_{ParaOnly+NoBoost}$ & clint paused on the trail. & \low{--}\\
            & & & \modelshort$_{ParaOnly+Boost}$ & clint stopped on the trail. & \high{+}\\
            & & & \modelshort$_{Joint+SupplyVerb}$ & clint walked on the trail. & \high{+}\\
            & & & \modelshort$_{Joint+NoBoost}$ & clint hiked on the trail. & =\\
            & & & \modelshort$_{Joint+Boost}$ & clint walked on the trail heading down. & \high{+}\\
       
    \bottomrule
    \end{tabular}
    \caption{Full version of Table \ref{tab:examples}.
    Example revisions from various models for sentences from the dev. set. Columns are: the target change in agency from the original to the target agency, the input sentence, the model, generated output, and the actual agency level of the output measured by the connotation frame lexicon.}
    \label{tab:more-examples}

\end{table*}

\section{Gender Bias in Movies}\label{sup:movie-bias}
\subsection{Extracting gender from characters}
The movie scripts mention characters in all caps, making it easy to identify and extract them.
We then cross reference the name (or, description for unnamed characters, e.g., ``the doorman'') with a list of gendered names\footnote{\url{http://www.cs.cmu.edu/Groups/AI/util/areas/nlp/corpora/names/0.html}} and gendered words (e.g., ``waitress,'' ``policeman,'' ``police woman'').
To allow for better rewriting using our model, we split the narratives into sentences \cite[using NLTK's sentence tokenizer][]{bird2009natural}, and assign each sentence to a character if their name appears in the sentence.

\end{document}